\documentclass[letterpaper, 10 pt, conference]{ieeeconf} 

\IEEEoverridecommandlockouts                              

\usepackage{censor}
\usepackage{graphicx}
\usepackage{booktabs}
\usepackage{amsmath}
\usepackage{tabularx}  
\usepackage{array}    
\usepackage{amssymb}
\usepackage{listings}
\usepackage{xcolor}
\usepackage{cuted}

\usepackage{fancyhdr}

\fancypagestyle{arxivnotice}{
    \fancyhf{} 
    \chead{\large\color{gray}This paper has been accepted for publication at the 
    
    IEEE/RSJ International Conference on Intelligent Robots and Systems (IROS), 2026.\copyright\ IEEE}
}

\title{\LARGE 
\vspace{5mm}
\textbf{AERMANI-VLM: Structured Prompting and Reasoning for Aerial Manipulation with Vision Language Models}
}

\author{%
{Sarthak~Mishra}$^{*1}$,~{Rishabh~Dev~Yadav}$^{*2}$,~{Avirup~Das}$^{2}$,~{Saksham~Gupta}$^{1}$,~{Wei~Pan}$^{3}$,~and~{Spandan~Roy}$^{1}$%
\thanks{This work is supported partly by ``Edge-AI-Di.Vision'' project from Qualcomm Technologies and partly by the `UASAT' project sponsored by MeITY, India. $(^{*})$ denotes equal contribution.}%
    \thanks{$^{1}$Robotics Research Center, IIIT Hyderabad, India.
    Emails: \texttt{\{sarthak.mishra, saksham.gupta\}@research.iiit.ac.in}, \texttt{spandan.roy@iiit.ac.in}}%
    \thanks{$^{2}$Department of Computer Science, University of Manchester, UK.
    Emails: \texttt{rishabh.yadav@postgrad.manchester.ac.uk}}%
    \thanks{$^{3}$Newcastle University, UK.
    Email: \texttt{wei.pan2@newcastle.ac.uk}}%
}

\usepackage{amsmath}
\usepackage{amssymb}    
\usepackage{booktabs}   

\usepackage{listings}
\lstdefinestyle{jsonsmall}{
    basicstyle=\ttfamily\footnotesize,
    breaklines=true,
    frame=single,
    backgroundcolor=\color{gray!5},
    numbers=none,
    xleftmargin=2pt,
    xrightmargin=2pt,
    aboveskip=2pt,
    belowskip=2pt
}

\usepackage{mathtools}
\usepackage{algorithm, algpseudocode}
\usepackage{graphicx, wrapfig, xcolor, multirow, cite}
\usepackage{url}
\usepackage{bbm}

\usepackage{booktabs}
\usepackage{array}
\algnewcommand{\Initialise}[1]{%
  \State \textbf{Initialise:}
  \Statex \hspace*{\algorithmicindent}\parbox[t]{.8\linewidth}{\raggedright #1}
}
\usepackage{amssymb, xcolor}

\begin{document}


\maketitle
\thispagestyle{arxivnotice}

\begin{abstract}
The rapid progress of vision–language models (VLMs) has sparked growing interest in robotic control, where natural language can express the operation goals while visual feedback links perception to action. However, directly deploying VLM-driven policies on aerial manipulators remains unsafe and unreliable since the generated actions are often inconsistent, hallucination-prone, and dynamically infeasible for flight.
In this work, we present AERMANI-VLM, the first framework to adapt pretrained VLMs for aerial manipulation by separating high-level reasoning from low-level control, without any task-specific fine-tuning. Our framework encodes natural language instructions, task context, and safety constraints into a structured prompt that guides the model to generate a step-by-step reasoning trace in natural language. This reasoning output is used to select from a predefined library of discrete, flight-safe skills, ensuring interpretable and temporally consistent execution.
By decoupling symbolic reasoning from physical action, AERMANI-VLM mitigates hallucinated commands and prevents unsafe behavior, enabling robust task completion. We validate the framework in both simulation and hardware on diverse multi-step pick-and-place tasks, demonstrating strong generalization to previously unseen commands, objects, and environment.

{\footnotesize
\noindent Website:  \url{https://sites.google.com/view/aermani-vlm}}
\end{abstract}

\section{Introduction}
Aerial manipulators (AMs) extend multirotors from passive sensing to active interaction, enabling grasping, transport, and placement of objects in cluttered or hard-to-reach environments \cite{ollero2021past, yadav2025integrated, yadav2024modular}. To be broadly useful, AMs should follow human-friendly instructions expressed in natural language, allowing non-expert operators to specify complex tasks without hand-crafted scripts.

Achieving vision-language-guided aerial manipulation remains difficult, because AMs must simultaneously interpret high-level natural language and execute tasks under the coupled demands of perception, navigation, and manipulation. Executing a command like “find the red cup” involves: (i) exploration, when the target is out of view; (ii) grounding, linking linguistic concepts (``red cup'') to visual observation; and (iii) execution, performing contact-rich motion respecting dynamics. These interdependent demands make AM's uniquely sensitive to perception or reasoning errors. Any reasoning failure directly compromises both grasping task success and the robot itself.

Existing Vision-Language-Action (VLA) models such as SayCan~\cite{saycan2022}, PaLM-E~\cite{palm-e2023}, RT-2~\cite{rt2_2023}, and OpenVLA~\cite{openvla2024} showcase strong zero-shot semantic reasoning for ground robots. However, directly transferring these systems to aerial manipulators is non-trivial. Unlike ground platforms, aerial robots operate under tight dynamic constraints, limited onboard computation, and safety-critical proximity to obstacles during grasping and placement. The data-intensive training and low-frequency inference typical of VLAs~\cite{Foehn_2022}, combined with brittle spatial grounding that can hallucinate unseen targets or misinterpret spatial relations~\cite{steiner2025mindmap}, and step-wise stochastic outputs~\cite{wu2025vulnerabilityllmvlmcontrolledrobotics}, conflict with the stability and control consistency required for flight. Moreover, many VLA systems generate context-free decisions without explicit reasoning over task history, leading to temporally inconsistent skill transitions. Similarly, Vision-Language Navigation (VLN) systems~\cite{huang2022uavvln} enable language-conditioned navigation for multirotors but do not extend to manipulation. Furthermore, the behavior of vision-language models (VLMs) is highly prompt-sensitive~\cite{gu2023systematic, zhou2022learning}, underscoring the need for structured decision mechanisms that enforce temporal coherence and execution consistency in aerial manipulation.

This gap raises a central question: \textit{how can general-purpose VLMs be adapted for aerial manipulation without costly domain training or fine-tuning?} To address these challenges, we propose \textbf{AERMANI-VLM}, a zero-shot framework that transfers pretrained vision–language reasoning models to aerial manipulation without task-specific fine-tuning.

The core idea is to decouple \textit{what to do} from \textit{how to do} it: the VLM handles high-level reasoning under structured prompting, while a predefined library of flight-safe skills executes the resulting actions. This design retains the generalization ability of large models while promoting temporally consistent and repeatable execution under aerial flight constraints.

\begin{figure*}[t]
\centering
\includegraphics[width=\textwidth]{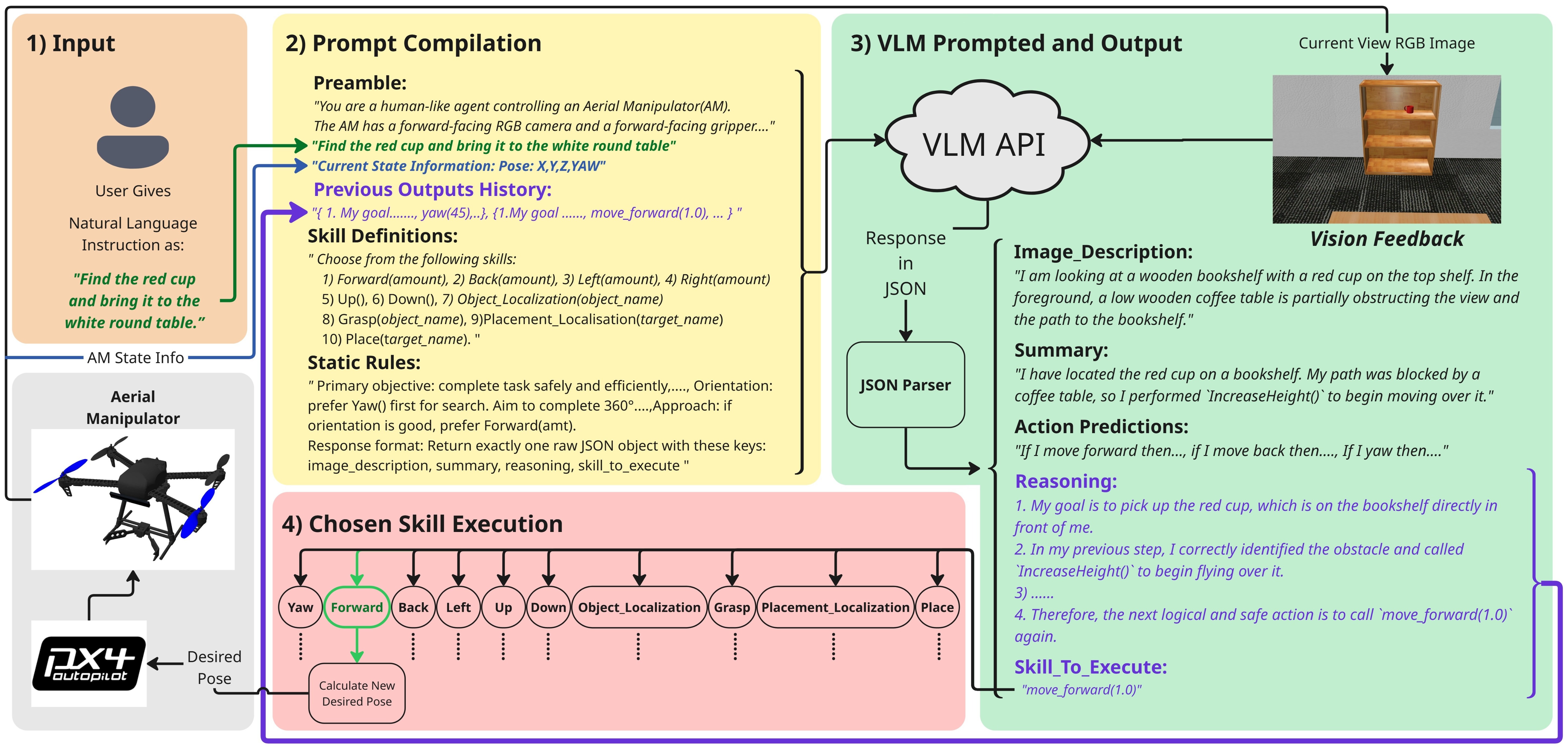}
\caption{Overview of the AERMANI-VLM pipeline for vision-language-guided aerial manipulation. 
(1) \textbf{Input:} a user provides a natural language command. 
(2) \textbf{Prompt compilation:} the command is compiled into a structured prompt containing a preamble, reasoning history, skill definitions, and safety rules. 
(3) \textbf{VLM inference:} together with the current RGB observation, the prompt is processed by a pretrained VLM, which outputs an image description, task summary, explicit reasoning trace, and a discrete skill to execute. 
(4) \textbf{Skill execution:} the selected motion primitive or perception-driven routine is executed by deterministic low-level controllers, ensuring repeatability under flight dynamics. 
This reasoning–action loop continues until task completion, enabling the VLM to focus on semantic reasoning while delegating precise execution to robust controllers.}
\label{fig:reasonfly_pipeline}
\vspace{-6pt}
\end{figure*}
The main contributions of this work are:
\begin{itemize}
    \item We introduce \textbf{AERMANI-VLM}, a framework for integrating pretrained vision–language models with aerial manipulation through a structured decision interface that separates high-level reasoning from flight-safe skill execution.

    \item We propose the \textbf{Descriptive Reasoning Trace (DRT)}, a structured reasoning representation that explicitly links visual grounding, task history, and candidate skill evaluation before action selection, improving temporal consistency in long-horizon manipulation tasks.

    \item We demonstrate that structured reasoning significantly improves decision stability and task success through controlled ablations and comparisons across multiple VLM backbones.

    \item We validate the framework in both simulation and real-world aerial manipulation experiments, showing reliable zero-shot execution of language-conditioned pick-and-place tasks.
\end{itemize}


\vspace{-1mm}

\subsection{Related Works}

Aerial manipulation has evolved from vision-guided systems using onboard cameras and artificial cues~\cite{Kim2016_visionGuidedAerial}, to markerless onboard grasping~\cite{Bauer2022_markerlessGrasp}, and end-effector-centric frameworks for versatile manipulation~\cite{he2025flying}. However, these works emphasize execution rather than language-level reasoning. Residual dynamics learning further reduces dependence on exact analytical models~\cite{cao2024computation,das2025dronediffusion,ujjawal2026learn,yadav2026learning,ujjawal2025aermani,yadav2025arcade,yadav2026physics}.

Vision--language--action models (VLAs)~\cite{saycan2022,palm-e2023,rt2_2023,openvla2024} combine LLM-based planning~\cite{brown2020gpt3,wei2022chain} with perceptual grounding through CLIP~\cite{radford2021clip}, CLIPort~\cite{shridhar2021cliport}, and LLaVA~\cite{liu2023llava}. Many use a language planner to select discrete skills executed by low-level controllers. Although effective for ground robots, they do not explicitly address aerial dynamics, safety-critical transitions, or structured evaluation of all candidate actions at each timestep. Modular approaches that separate reasoning from skill execution~\cite{hu2023toward,zhang2025safevlasafetyalignmentvisionlanguageaction} improve executability but remain primarily validated in ground manipulation and navigation. DroneVLA~\cite{sun2026airvlavisionlanguageactionsystemsaerial} and AIR-VLA~\cite{mehboob2026dronevlavlabasedaerial} extend language reasoning to aerial platforms, but focus mainly on navigation and perception rather than manipulation.

For multirotors, foundation-model research has addressed mission planning~\cite{sautenkov2025uavvla, song2025soranav}, spatial reasoning~\cite{gao2025openfly}, direct control~\cite{lykov2025cognitivedrone}, and broader agentic UAV autonomy requiring exploration, grasping, and placement~\cite{tian2025uavs}. Language and vision models have also been applied to aerial grasping and placement~\cite{singh2026aerograb,mishra2026aeroplace}.

Thus, existing work separately addresses aerial execution, semantic reasoning, or flight autonomy. A unified framework that constrains vision--language reasoning within a structured, temporally consistent decision loop for aerial manipulation remains underexplored. AERMANI-VLM extends the planner--skill paradigm with explicit decision constraints that promote coherent skill transitions under aerial dynamics.

\section{AERMANI-VLM: Methodology}
\begin{figure}[t]
\centering
\includegraphics[width=0.35\textwidth]{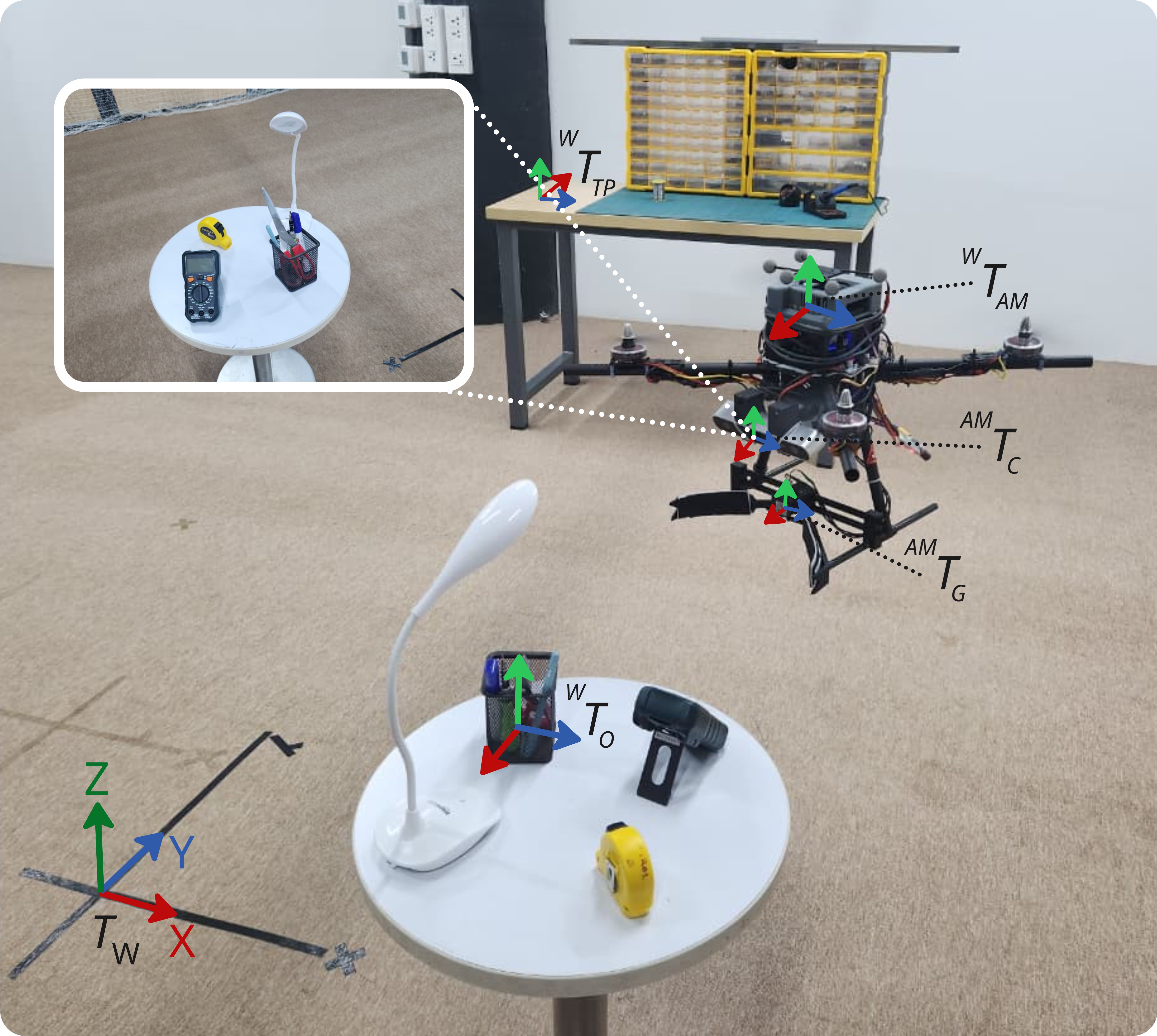}
\caption{Coordinate frames and spatial grounding in {AERMANI-VLM}. 
\textbf{(i)} The global world frame $T_W$ anchors all transformations, defining poses for the aerial manipulator ($^{W}T_{AM}$), target object ($^{W}T_O$), and placement location ($^{W}T_{TP}$). 
\textbf{(ii)} Onboard frames for the camera ($^{AM}T_C$) and gripper ($^{AM}T_G$) are expressed relative to the manipulator body, maintaining consistency between perception and control. 
}
\label{fig:frames}
\vspace{-15pt}
\end{figure}

This section introduces its core components: problem formulation, prompt design, VLM output structure, the skill library, and the closed reasoning–action loop. 
Table~\ref{tab:notations} summarizes the key notations, and Fig.~\ref{fig:frames} illustrates the coordinate frames used throughout. 

\begin{table}[h]
\centering
\caption{Notation summary for AERMANI-VLM.}
\label{tab:notations}
\begin{tabular}{ll}
\toprule
Symbol & Description \\
\midrule
$L$ & Natural language command \\
$P_t$ & Structured prompt at timestep $t$ \\
$s_t$ & Latent world state (partially observable) \\
$o_t$ & Onboard observation at timestep $t$ \\
$a_t \in A$ & Skill action from skill library $A$ \\
$\pi$ & High-level VLM policy \\
$\gamma_t$ & Structured VLM output at timestep $t$ \\
\bottomrule
\end{tabular}
\vspace{-5mm}
\end{table}

\subsection{Problem Formulation and Architecture Overview}

We consider language-guided aerial manipulation as a sequential perception–reasoning–action process. A user specifies a task using a natural language command $L$ (e.g., ``pick up the red cup and place it on the table''), and the aerial manipulator must interpret this instruction and execute a sequence of actions using only onboard perception.

At each timestep $t$, the robot observes the environment through onboard sensors and must decide the next high-level action required to progress toward task completion. The true world state, including object poses and placement targets, is not fully observable and must be inferred from visual observations over time. Formally, the robot receives an observation $o_t = \{I_t, q_t\}$, where $I_t$ is the current RGB-D image from the onboard camera and $q_t$ represents the multirotor pose (position and yaw). A pretrained vision–language model (VLM) is used as a high-level reasoning policy. Given the current observation $o_t$ and a structured task prompt $P_t$, the VLM predicts the next high-level decision: $\gamma_t = \pi(o_t, P_t).$

The output $\gamma_t$ is a structured response containing two components:
\begin{itemize}
\item a reasoning trace describing the model's interpretation of the scene and task context, and
\item a discrete skill identifier corresponding to an executable action.
\end{itemize}

The selected skill is executed by the robot using predefined low-level controllers. After execution, the resulting observations and reasoning outputs are appended to the task history and incorporated into the next prompt. This process forms a closed perception–reasoning–action loop:
\begin{equation}
L
;\xrightarrow{\text{task}}
P_t
;\xrightarrow{\text{query}}
\pi(o_t, P_t)
;\xrightarrow{\text{VLM output}}
\gamma_t
;\xrightarrow{\text{execution}}
a_t \nonumber
\end{equation}
Here, the VLM is responsible for high-level reasoning and task planning, while the physical execution of actions is handled by a library of predefined, flight-safe skills. This separation allows the system to leverage the semantic reasoning capabilities of large vision–language models while maintaining reliable and dynamically feasible control for aerial manipulation.

\subsection{Structured Prompt Design and Input Encoding}
To guide high-level decision making, the natural language command and task context are compiled into a structured prompt $P_t$ before querying the VLM. The prompt provides the model with the necessary task description, operational constraints, and execution history required for selecting the next  and consists of 5 main parts:
\begin{equation}
P_t = \{L,~\textit{Preamble},~\textit{History},~\textit{Skill Library},~\textit{Rules}\}.
\end{equation}

\textbf{Task Command ($L$).} The user-provided natural language instruction defining the task objective.
\textbf{Preamble.} A short description of the robotic platform, sensing configuration, and the role of the VLM within the control loop.
\textbf{History.} A structured record of previously executed skills together with their termination status and key perceptual outcomes. This memory allows the VLM to reason over past decisions and their observed consequences, helping maintain temporal consistency and avoid repeated or contradictory actions.
\textbf{Skill Library.} A description of the available motion and perception skills that define the discrete action space of the aerial manipulator.
\textbf{Rules.} Output formatting constraints that require the model to produce a structured response consisting of a reasoning trace (\textit{DRT}) and a selected skill to execute (\textit{STE}).

\subsection{Structured Output for Executable Reasoning}
\label{sect:2C}

Given the observation $o_t$ and structured prompt $P_t$, the pretrained VLM produces a bounded and machine-interpretable response:
\begin{equation}
\gamma_t = \pi(o_t, P_t) = \{\textit{DRT}_t,~\textit{STE}_t\}.
\end{equation}

The output consists of two components: a structured reasoning trace and a discrete skill selection. The reasoning trace, referred to as the \textit{Descriptive Reasoning Trace (DRT)}, organizes the model’s perception, memory, and action evaluation into four fields:
\begin{equation}
\begin{split}
\mathrm{DRT}_t = \{
& \mathrm{Image~Description},~
\mathrm{Summary},~
\\
& \mathrm{Action~Predictions},~
\mathrm{Reasoning}
\}.
\end{split}
\end{equation}

\textbf{Image Description.} Describes the current visual scene, including visible objects, their spatial arrangement, and other contextual cues. This grounds the reasoning process in the current observation.

\textbf{Summary.} Provides a short recap of previous observations and executed skills, enabling the model to maintain awareness of task progress across reasoning steps.

\textbf{Action Predictions.} Enumerates the available skills and briefly estimates the outcome of executing each candidate action. This encourages the model to evaluate the full action set before selecting the next skill.

\textbf{Reasoning.} Integrates the current observation, task history, and predicted outcomes to justify the final action choice.
\indent The second component of the output, $\textit{STE}_t$, specifies the \textit{Skill To Execute}, which corresponds to a discrete action $a_t \in A$ selected from the predefined skill library. The DRT therefore records the reasoning process leading to the selected action, while the STE provides the executable command for the robot.

This output format constrains the VLM to produce interpretable, verifiable reasoning that links visual perception to an executable action, forming a transparent reasoning–execution bridge for aerial manipulation.

\subsection{Skill Library for Aerial Manipulation} \label{sect:2C}

The skill library $A$ defines the discrete set of executable routines available to the aerial manipulator.  
The VLM selects one skill, which is executed as a predefined routine with fixed action sequences and termination conditions.
The library comprises two categories:  

\textbf{1) Motion Primitives.}  
The motion primitives are user-defined pose adjustments of the aerial manipulator in its body frame, enabling active exploration, re-alignment, and precise positioning when targets object or placement sites are not visible. Each primitive is executed as a predefined routine with fixed angle and distance parameters, ensuring consistent behavior across tasks. 
The motion primitives are summarized in Table~\ref{tab:motion_primitives}.

\begin{table}[h]
\vspace{-3mm}
\centering
\caption{Motion primitives in the skill library.}
\label{tab:motion_primitives}
\begin{tabular}{p{2.5cm} p{5.0cm}}
\toprule
Primitive & Description \\
\midrule
\texttt{yaw} & Rotate about the body $z$-axis for view re-alignment or orientation adjustment. \\
\texttt{forward/backward} & Translate along the body $x$-axis to approach or retreat from detected targets. \\
\texttt{left/right} & Translate along the body $y$-axis to correct lateral misalignment or bypass occlusions. \\
\texttt{up/down} & Adjust altitude to maintain clearance or reach manipulation height. \\
\bottomrule
\end{tabular}
\end{table}

\textbf{2) Perception-Driven Manipulation Routines.} Composite routines that couple perception with control for physical interaction. The four core routines are summarized in Table~\ref{tab:perception_routines}.  

\begin{table}[h]
\centering
\caption{Perception-driven routines in the skill library.}
\label{tab:perception_routines}
\begin{tabular}{p{1.0cm} p{6.5cm}}
\toprule
Routine & Description \\
\midrule
\texttt{obj\_localization} & ~~~~~~~~~~~~~~Estimate 3D pose of the queried object using SAM3~\cite{carion2025sam3segmentconcepts} for zero-shot segmentation fused with depth and camera extrinsics. \\
\texttt{grasp} & Iterative re-localization with yaw alignment, closed-loop gripper actuation, and safe retreat after contact. \\
\texttt{placement\_localization} & ~~~~~~~~~~~~~~~~~~~~~~~~Identify a candidate surface in the point cloud and compute a collision-free placement pose with vertical offset. \\
\texttt{place} & Controlled descent, release, and retreat to a safe hover. \\
\bottomrule
\end{tabular}
\vspace{-5mm}
\end{table}

All motion and perception routines are predefined, yielding repeatable behavior under identical high-level inputs (e.g., move left, ascend, rotate).
Each skill follows a fixed control–perception sequence, decoupling the VLM’s reasoning from low-level dynamics (e.g., thrust, body rates).
This modular separation enables new skills to be added or refined without modifying the reasoning framework or compromising system safety.

\subsection{Reasoning–Action Execution Loop}
The reasoning–action execution loop forms the closed control cycle that links perception, reasoning, and actuation. At each timestep $t$, the aerial manipulator first acquires the current observation $o_t$ and constructs the structured prompt $P_t$ using the task command, execution history, and skill definitions. The pretrained VLM is then queried to produce the structured output $\gamma_t = \{\textit{DRT}_t, \textit{STE}_t\}.$

The selected skill $\textit{STE}_t$ is executed as a predefined routine until its termination condition is reached. After execution, a compact summary of the outcome—including the executed skill, its termination status (e.g., success, failure, or timeout), and relevant perceptual indicators—is appended to the History buffer together with the corresponding reasoning trace. This updated history is incorporated into the next prompt, allowing subsequent decisions to be conditioned on both prior reasoning and observed execution results. For example, the system may trigger re-localization after a failed grasp attempt or resume exploration if the target object is no longer visible.

The system automatically re-prompts the VLM if the generated output is invalid or outside the skill library. Additionally, a lightweight working memory stores detected objects and executed actions, helping prevent redundant exploration and supporting consistent decision making across reasoning steps. This perception–reasoning–execution loop continues until the task is completed. The overall process is summarized in Algorithm~\ref{alg:reasonfly} and illustrated in Fig.~\ref{fig:reasonfly_pipeline}.

\begin{algorithm}[t]
\caption{AERMANI-VLM reasoning–action loop.}
\label{alg:reasonfly}
\begin{algorithmic}[1]
\Require Natural language command $L$, skill library $A$
\Ensure Executed skill sequence $\{a_1, a_2, \dots, a_T\}$
\State Initialize history buffer $H \gets \emptyset$
\For{each timestep $t = 1,2,\dots$ until task completion}
    \State Acquire observation: $o_t = \{I_t, q_t\}$
    \State Construct structured prompt:
    \[
    P_t = \{L, \text{Preamble}, H, \text{Skill Library}, \text{Rules}\}
    \]
    \State Query VLM policy:
    \[
    \gamma_t = \pi(o_t, P_t) = \{\text{DRT}_t, \text{STE}_t\}
    \]
    \State Extract skill: $a_t \gets \text{STE}_t \in A$
    \State Execute $a_t$ as a predefined routine
    \State Update history: $H \gets H \cup \{\text{DRT}_t, a_t\}$
\EndFor
\end{algorithmic}
\end{algorithm}

\section{Experiments}

Experiments are conducted in both simulation and real-world settings to evaluate the performance of AERMANI-VLM and compare it with existing approaches.  To assess the generality of the proposed framework across different vision–language backbones, we evaluate it using multiple pretrained VLMs, including \textit{Gemini~3-Pro}, \textit{GPT-4o}, \textit{Claude~3.5 Sonnet}, \textit{Qwen2-VL-7B}, and \textit{LLaVA-Next-13B}. All models are used in a zero-shot setting without any task-specific fine-tuning. For perception, we employ SAM3~\cite{carion2025sam3segmentconcepts} for open-vocabulary segmentation and mask extraction. Each policy operates by selecting a single discrete skill from the predefined skill library at every reasoning step.

\subsection{Experimental Setup}  
\begin{figure}[t]
\centering
\includegraphics[width=0.48\textwidth]{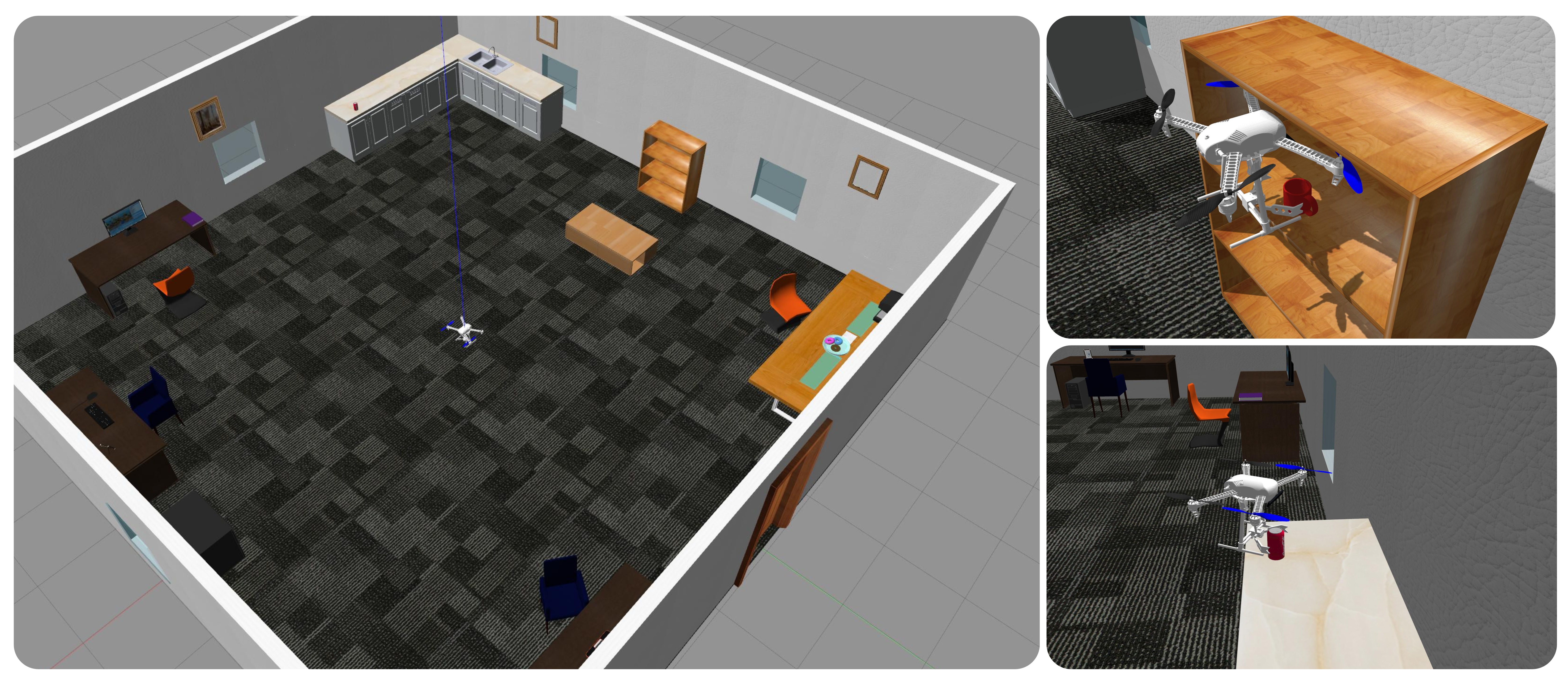}
\caption{Simulation Environment consisting of a custom indoor office setup with randomized layouts of tables, shelves, and other furniture.
}
\label{fig:sim_environ}
\vspace{-12pt}
\end{figure}

\noindent \textbf{Simulation:}  
The AM is operated wiith PX4 Autopilot in a Software-in-the-Loop (SITL) configuration, enabling realistic flight control. A custom manipulator was mounted on the simulated \textit{Iris} quadrotor to replicate the real-world hardware design.
\noindent  \textbf{Hardware:} The physical platform was a Tarot~650 hexarotor equipped with a CUAV~X7+ flight controller running PX4. High-level autonomy and perception ran on an NVIDIA Jetson Orin Nano Super using ROS, with onboard sensing provided by a ZED RGB-D camera and a two-finger servo gripper. The Jetson computed skill-level setpoints, while the PX4 executed low-level way-point tracking. he experimental environment in simulation is shown in Fig~\ref{fig:sim_environ}. 

\subsection{Task Suites}

We design four progressively challenging task categories that test complementary aspects of language-conditioned aerial manipulation.  
\textbf{(i) Localization and Identification} tasks involve clear, unambiguous object–language references (e.g., “Pick up the red cup and place it on the table”), assessing perception–language grounding.  
\textbf{(ii) Pick from Clutter / Ambiguity Resolution} introduces multiple candidate objects (e.g., “Bring me a snack” when both an apple and a cracker box are visible), probing reasoning under perceptual and linguistic ambiguity.  
\textbf{(iii) Sequential Instruction Following} tasks combine navigation, perception, and manipulation (e.g., “Move closer to the table, turn left, and pick up the cup”), evaluating sequential consistency and procedural reasoning.  
Finally, \textbf{(iv) Search and Pick} requires locating targets initially outside the field of view (e.g., “Find the soda can on the table and bring it here”), testing exploratory reasoning before manipulation.  
Representative commands for all categories are listed in Table~\ref{tab:commands_list_full}. 

\begin{table}[h!]
\centering
\caption{\small Representative task commands.}
\vspace{-8pt}
\label{tab:commands_list_full}
\begin{tabular}{p{0.9\columnwidth}}
\toprule
\textbf{Task and Example Commands} \\
\midrule
\textbf{Localization \& Identification} \\
``Pick up the [Object] and place it on the [Location].'' \\
``Get the [Object] and put it in the [Location].'' \\
(Objects: red cup, coke can, tissue box; Locations: table, sink, shelf). \\
\addlinespace[0.6em]

\textbf{Pick from Clutter / Ambiguity Resolution} \\
``Of the two boxes, pick up the larger one.'' \\
``Ignore the tools and pick up the green bottle.'' \\
\addlinespace[0.6em]

\textbf{Sequential Instruction Following} \\
``Move forward twice, turn right, and pick up the tissue box.'' \\
``Approach the counter. If you see an orange, pick it up; otherwise grab the banana.'' \\
\addlinespace[0.6em]

\textbf{Search and Pick} \\
``Find the purple cup and place it on the round table.'' \\
``Locate the cardboard box and bring it to the sink.'' \\
\bottomrule
\end{tabular}
\vspace{-8pt}
\end{table}

\begin{figure*}[t]
\centering
\includegraphics[width=0.85\textwidth]{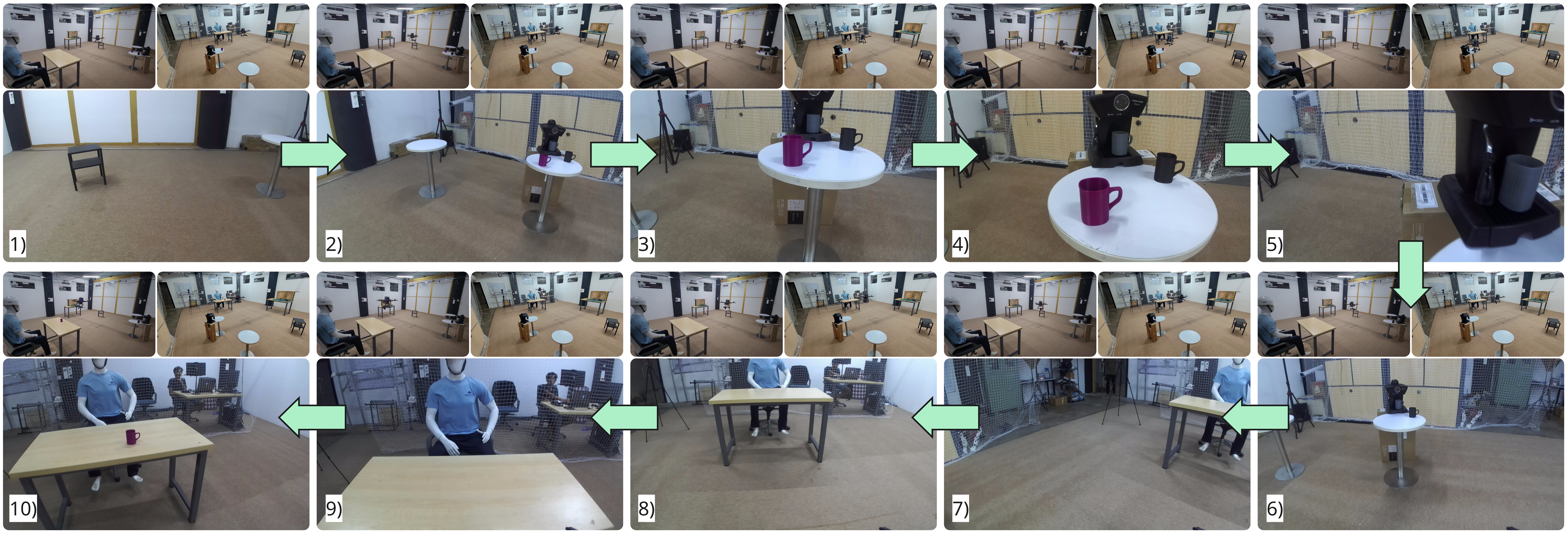}
\caption{Qualitative results from a real-world hardware experiment for the command: ``Pick up the purple cup next to the coffee machine and place it on the wooden table.'' Each panel shows the first-person onboard view (large bottom image) and two static third-person views (small top images). The numbered sequence illustrates the complete, autonomous execution of the task. {(1--3)} The AM performs an active search to find the target object. {(4--5)} It executes a visually-guided approach and grasp. {(6--8)} After securing the cup, it searches for the destination table. {(9--10)} Finally, it approaches the table and places the object. }
\label{fig:hardware_exps}
\vspace{-10pt}
\end{figure*}

\begin{figure*}[t]
\centering
\includegraphics[width=0.85\textwidth]{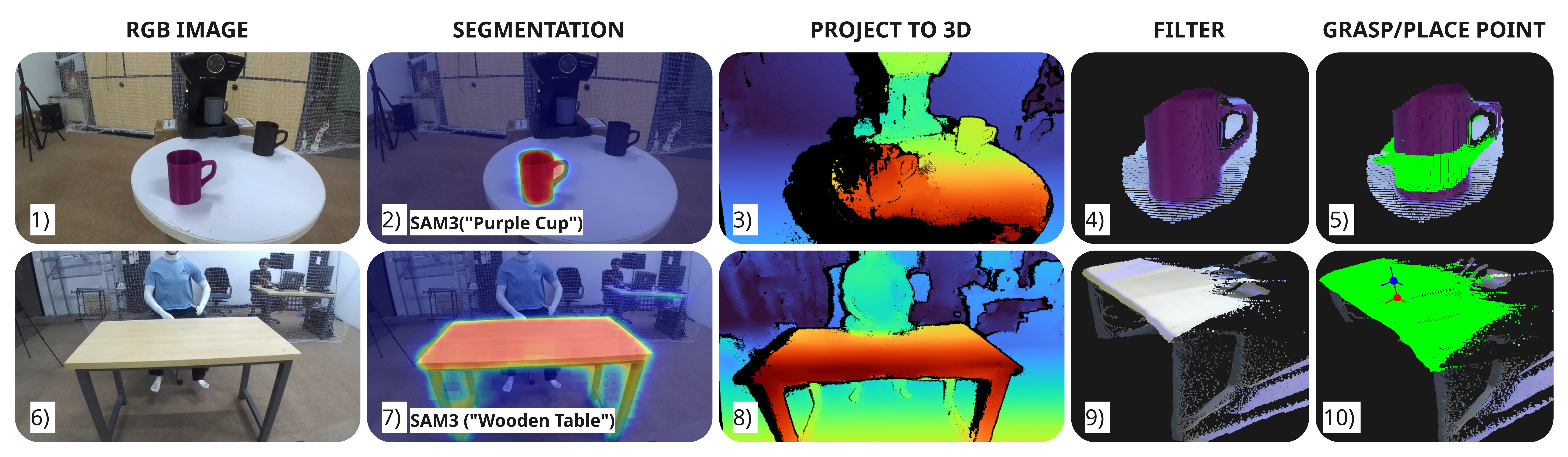}
\caption{The open-vocabulary perception pipeline for object\_localization (top row) and placement\_localization (bottom row). Given a natural language query (e.g., "Purple Cup"), SAM3 performs zero-shot segmentation on the input RGB image. This 2D mask is then used to extract a filtered 3D point cloud from the depth data, allowing for the precise calculation of a grasp or placement pose.}
\label{fig:clipseg}
\vspace{-12pt}
\end{figure*}

\subsection{Evaluation Metrics}

Performance is evaluated using complementary task-level, reasoning-level, and system-level metrics. All metrics are computed separately for each VLM backbone across multiple trials.
\textbf{Task Success Rate (SR).} Percentage of trials in which the robot successfully completes the instructed task, defined as 
$\text{SR} = \frac{N_{\text{success}}}{N_{\text{total}}}\times100$. 
A trial is successful if the correct object is grasped and placed within a spatial tolerance before the timeout $T_{\max}=10\,\text{min}$.
\textbf{Action Redundancy Ratio (ARR).} Measures unnecessary or recoverable actions during successful trials, computed as 
$\text{ARR} = \frac{1}{N_{\text{success}}}\sum_i \frac{n^{(i)}_{\text{redundant}}}{n^{(i)}_{\text{total}}}$. 
Redundant actions include oscillatory corrections or zero-net-displacement motions identified from trajectory logs.
\textbf{Reasoning Steps (RS).} Average number of VLM reasoning cycles per task, defined as 
$\text{RS} = \frac{1}{N_{\text{total}}}\sum_i n^{(i)}_{\text{VLM}}$.
\textbf{Invalid Action Rate (IAR).} Percentage of VLM outputs that do not correspond to a valid skill and require re-prompting, defined as 
$\text{IAR} = \frac{N_{\text{invalid}}}{N_{\text{queries}}}\times100$.
\textbf{VLM Response Time (RT).} Median and 95th-percentile inference latency per VLM query.
\textbf{End-to-End Task Time (TT).} Total execution time required to complete successful tasks.

\subsection{Model and Baseline Comparison}

We compare AERMANI-VLM with representative vision--language and aerial reasoning frameworks across multiple VLM backbones. All methods receive identical RGB observations and language commands and use the same discrete aerial skill library through the PX4 control pipeline, ensuring that differences reflect high-level decision policies rather than low-level perception or control.

The baselines represent complementary paradigms. \textbf{RT-2}~\cite{rt2_2023} directly predicts discrete actions from multimodal inputs, while \textbf{SayCan}~\cite{saycan2022} selects skills through language-based affordance grounding. \textbf{UAV-VLN}~\cite{huang2022uavvln} predicts language-conditioned aerial waypoints, whereas \textbf{UAV-VLA/CognitiveDrone}~\cite{lykov2025cognitivedrone,sautenkov2025uavvla} performs high-level reasoning for UAV missions. Because these methods were designed primarily for navigation or ground manipulation, their outputs are mapped to the shared aerial skill interface without retraining.

We also include a \textbf{ReAct-style prompting baseline} using the same VLM backbone, with chain-of-thought reasoning followed by JSON-formatted skill selection. This isolates the benefit of the proposed Descriptive Reasoning Trace (DRT) beyond generic structured prompting.

Table~\ref{tab:model_baseline_comparison} shows that RT-2 lacks temporal stability, SayCan often generates redundant skill sequences, and UAV-VLN/UAV-VLA lack manipulation primitives or fine-grained control. ReAct improves reasoning stability but does not explicitly evaluate candidate actions or enforce temporal grounding. AERMANI-VLM achieves the highest success rate and lowest action redundancy through structured prompting, DRT-based reasoning, and deterministic skill execution.

We further evaluate AERMANI-VLM using \textit{Gemini~3-Pro}, \textit{GPT-4o}, \textit{Claude~3.5 Sonnet}, \textit{Qwen2-VL-7B}, and \textit{LLaVA-Next-13B}. Results remain consistent across backbones, with frontier models showing slightly higher success and lower reasoning overhead. This confirms that the framework generalizes across VLM architectures without task-specific fine-tuning.

\begin{table}[t]
\footnotesize
\renewcommand{\arraystretch}{1.1}
\caption{\small Comparison with baseline frameworks and different VLM backbones. Metrics averaged across simulation tasks.}
\centering
\scalebox{0.75}{
\begin{tabular}{lcccccc}
\toprule
\textbf{Method} & \textbf{SR$\uparrow$} & \textbf{ARR$\downarrow$} & \textbf{RS$\downarrow$} & \textbf{IAR$\downarrow$} & \textbf{RT(s)$\downarrow$} & \textbf{Manip.} \\
\midrule

RT-2~\cite{rt2_2023} & 10.0 & 19.0 & 42.0 & 35.0 & 0.85 & \checkmark \\
SayCan~\cite{saycan2022} & 20.0 & 16.0 & 33.0 & 28.0 & 0.91 & \checkmark \\
ReAct Prompting (VLM) & 35.0 & 12.5 & 28.0 & 15.0 & 0.90 & \checkmark \\
UAV-VLN~\cite{huang2022uavvln} & 30.0 & 16.5 & 38.0 & 31.0 & 0.72 & -- \\
UAV-VLA~\cite{lykov2025cognitivedrone,sautenkov2025uavvla} & 45.0 & 10.5 & 26.0 & 18.0 & 0.88 & -- \\

\midrule

\textbf{AERMANI-VLM (Gemini-3-Pro)} & \textbf{87.5} & \textbf{2.3} & \textbf{16.8} & \textbf{4.1} & 0.92 & \checkmark \\
AERMANI-VLM (GPT-4o) & 85.0 & 2.6 & 18.4 & 5.2 & 0.88 & \checkmark \\
AERMANI-VLM (Claude-3.5) & 82.5 & 3.1 & 19.6 & 6.4 & 0.95 & \checkmark \\
AERMANI-VLM (Qwen2-VL) & 77.5 & 4.5 & 22.8 & 8.7 & 0.61 & \checkmark \\
AERMANI-VLM (LLaVA-Next) & 72.5 & 5.2 & 25.3 & 10.1 & 0.58 & \checkmark \\

\bottomrule
\end{tabular}
}
\label{tab:model_baseline_comparison}
\vspace{-10pt}
\end{table}

\subsection{Ablation Studies}
\label{sec:ablation}

To isolate the contributions of structured prompting and the Descriptive Reasoning Trace (DRT), we perform controlled ablations while keeping the VLM backbone, perception pipeline, and skill library fixed. This ensures that performance differences arise from the reasoning interface rather than model capacity.

\begin{table}[h!]
\centering
\caption{Ablation Configuration}
\vspace{-6pt}
\scalebox{0.87}{
\begin{tabular}{lccccccccc}
\toprule
\multirow{2}{*}{\textbf{Variant}} & \multicolumn{5}{c}{\textbf{Prompt Components}} & \multicolumn{4}{c}{\textbf{DRT Components}} \\
\cmidrule(lr){2-6} \cmidrule(lr){7-10}
 & L & P & H & S & $R^{*}$ & I & Su & A & Re \\
\midrule
Unstructured Prompting  & $\checkmark$ & $\times$ & $\times$ & $\checkmark$ & $\times$ & $\times$ & $\times$ & $\times$ & $\times$ \\
No-DRT (Direct Control) & $\checkmark$ & $\checkmark$ & $\checkmark$ & $\checkmark$ & $\checkmark$ & $\times$ & $\times$ & $\times$ & $\times$ \\
Reduced DRT--V1 & $\checkmark$ & $\checkmark$ & $\checkmark$ & $\checkmark$ & $\checkmark$ & $\checkmark$ & $\times$ & $\times$ & $\times$ \\
Reduced DRT--V2 & $\checkmark$ & $\checkmark$ & $\checkmark$ & $\checkmark$ & $\checkmark$ & $\checkmark$ & $\checkmark$ & $\times$ & $\times$ \\
Reduced DRT--V3 & $\checkmark$ & $\checkmark$ & $\checkmark$ & $\checkmark$ & $\checkmark$ & $\checkmark$ & $\checkmark$ & $\checkmark$ & $\times$ \\
Full AERMANI-VLM & $\checkmark$ & $\checkmark$ & $\checkmark$ & $\checkmark$ & $\checkmark$ & $\checkmark$ & $\checkmark$ & $\checkmark$ & $\checkmark$ \\
\bottomrule 
\end{tabular}
}
\vspace{2pt}

\parbox{0.95\columnwidth}{\footnotesize
\textbf{Component definitions:}  
L = Language Command,  
P = Preamble,  
H = History,  
S = Skill Library,  
R = Rules,  
I = Image Description,  
Su = Summary,  
A = Action Predictions,  
Re = Reasoning.  
{*} variant-specific modification.}
\label{tab:variant_config}
\vspace{-4pt}
\end{table}

\subsubsection{Ablation Setup}

We evaluate six variants of the system, summarized in Table~\ref{tab:variant_config}. The variants progressively introduce structured prompt components and DRT fields.

\begin{itemize}
\item \textbf{Unstructured Prompting}: Minimal prompt containing only the task command and skill list, without structural constraints.
\item \textbf{No-DRT}: Uses the full structured prompt but the model directly outputs the selected skill without an explicit reasoning trace.
\item \textbf{DRT-V1}: Adds the \emph{Image Description} field to ground reasoning in the current observation.
\item \textbf{DRT-V2}: Introduces the \emph{Summary} field, enabling short-term memory of past actions.
\item \textbf{DRT-V3}: Adds \emph{Action Predictions}, allowing the model to evaluate candidate skills before selection.
\item \textbf{Full AERMANI-VLM}: Uses the complete DRT representation including the final reasoning field.
\end{itemize}

This sequence allows us to evaluate how each reasoning component contributes to task performance and decision stability.

\begin{table}[t]
\footnotesize
\renewcommand{\arraystretch}{1.2}
\caption{\small Simulation results across four task categories. Metrics: Success Rate, Action Redundancy Ratio, Reasoning Steps, and Invalid Action Rate.}
\centering
\scalebox{0.80}{
\begin{tabular}{|c|l|c|c|c|c|}
\hline
\textbf{Task Suite} & \textbf{Method} & \textbf{SR~$\uparrow$} & \textbf{ARR~$\downarrow$} & \textbf{RS~$\downarrow$} & \textbf{IAR~$\downarrow$} \\ 
\hline

\multirow{6}{*}{Localization} 
& Unstructured & 10.0 & 21.0 & 40.0 & 34.5 \\ 
& No-DRT & 20.0 & 20.3 & 33.5 & 27.8 \\ 
& DRT-V1 & 30.0 & 18.1 & 31.4 & 22.6 \\ 
& DRT-V2 & 50.0 & 14.7 & 26.3 & 17.2 \\ 
& DRT-V3 & 60.0 & 10.2 & 21.5 & 11.4 \\ 
& \textbf{AERMANI-VLM} & \textbf{90.0} & \textbf{2.1} & \textbf{11.0} & \textbf{3.2} \\ 
\hline

\multirow{6}{*}{Pick from Clutter} 
& Unstructured & 0.0 & -- & 35.0 & 38.7 \\ 
& No-DRT & 10.0 & 17.1 & 28.0 & 30.4 \\ 
& DRT-V1 & 20.0 & 14.9 & 26.1 & 25.7 \\ 
& DRT-V2 & 40.0 & 12.6 & 22.5 & 18.9 \\ 
& DRT-V3 & 50.0 & 8.9 & 18.7 & 12.6 \\ 
& \textbf{AERMANI-VLM} & \textbf{80.0} & \textbf{4.0} & \textbf{14.1} & \textbf{5.3} \\ 
\hline

\multirow{6}{*}{Sequential} 
& Unstructured & 10.0 & -- & 42.0 & 36.2 \\ 
& No-DRT & 20.0 & 15.0 & 35.7 & 28.9 \\ 
& DRT-V1 & 30.0 & 12.1 & 32.4 & 24.3 \\ 
& DRT-V2 & 50.0 & 8.7 & 28.3 & 17.5 \\ 
& DRT-V3 & 60.0 & 6.5 & 25.4 & 12.8 \\ 
& \textbf{AERMANI-VLM} & \textbf{100.0} & \textbf{2.8} & \textbf{19.2} & \textbf{4.1} \\ 
\hline

\multirow{6}{*}{Search \& Pick} 
& Unstructured & 0.0 & 16.0 & 46.0 & 41.0 \\ 
& No-DRT & 10.0 & 14.3 & 39.8 & 33.7 \\ 
& DRT-V1 & 30.0 & 12.4 & 37.1 & 27.2 \\ 
& DRT-V2 & 50.0 & 9.5 & 33.6 & 20.1 \\ 
& DRT-V3 & 60.0 & 6.9 & 29.2 & 14.6 \\ 
& \textbf{AERMANI-VLM} & \textbf{80.0} & \textbf{3.1} & \textbf{23.7} & \textbf{6.2} \\ 
\hline

\end{tabular}
}
\label{tab:main_results_by_category}
\vspace{-10pt}
\end{table}

\subsubsection{Results}

Table~\ref{tab:main_results_by_category} summarizes simulation results across four task categories, with ten repetitions per configuration. Performance improves consistently as structured reasoning components are added. Full AERMANI-VLM achieves an average success rate of \textbf{87.5\%}, substantially exceeding \textbf{No-DRT (15.0\%)} and \textbf{Unstructured Prompting (5.0\%)}. It also reduces Action Redundancy Ratio (ARR) and Reasoning Steps (RS), indicating more stable and efficient decisions across all categories. Structured prompting improves language grounding, while successive DRT components enhance temporal consistency and reduce redundant exploration.

\subsubsection{Analysis}

The ablation confirms that structured reasoning is essential for stable aerial manipulation. \textbf{Unstructured Prompting} lacks explicit constraints and temporal context, producing inconsistent action sequences. \textbf{No-DRT} improves grounding but remains reactive, often repeating or reversing actions.
Adding DRT components progressively improves performance: \textbf{Image Description} strengthens visual grounding, \textbf{Summary} preserves task progress, and \textbf{Action Predictions} encourages evaluation of candidate skills before execution. The full AERMANI-VLM combines these elements into a consistent perception--memory--prediction loop, yielding the highest success rate and most efficient action sequences.

\subsection{Real-World Hardware Experiments}

We deployed the complete AERMANI-VLM system on the physical aerial manipulation platform described in Sec.~IV-A. Experiments were conducted indoors among tables, shelves, and household objects across \textbf{25 trials} using representative task commands. Each trial required interpreting a natural-language instruction, localizing the target through open-vocabulary perception, and completing a pick-and-place sequence.
AERMANI-VLM achieved an \textbf{82\%} task success rate, averaging \textbf{17.2 reasoning steps} and an action redundancy ratio of \textbf{2.3} per successful task. The median VLM response time was \textbf{0.92\,s} per query, supporting stable closed-loop execution. Qualitative results are shown in Fig.~\ref{fig:hardware_exps}. These experiments demonstrate reliable translation of language commands into executable aerial manipulation under real-world sensing and control constraints.

\vspace{-1mm}

\section{Conclusion and Future Work}
This work presented AERMANI-VLM, the first framework that systematically integrates pretrained VLM with aerial manipulation through structured prompting and predefined skill library. By decoupling semantic reasoning from physical control, the system transforms general-purpose VLMs into reliable decision layers for aerial robots without domain-specific retraining.
Extensive simulation and hardware experiments demonstrate that structured input–output design is critical: prompting defines the task boundary, while the output reasoning trace enforces temporal grounding and verifiable reasoning. Together, these mechanisms reduce hallucinations, stabilize task sequences, and enable zero-shot manipulation from natural-language commands.
Future work will extend this foundation toward dexterous and dual-arm manipulation, persistent long-horizon missions with memory, and collaborative multi-agent reasoning, advancing the vision of autonomous aerial systems that understand, reason, and act through language.

\vspace{-2mm}
\section*{ACKNOWLEDGMENT}
The authors acknowledge the use of LLMs for improving the grammatical structure and clarity of this manuscript.
\vspace{-2mm}

\bibliographystyle{IEEEtran}
\bibliography{final_bib} 

\end{document}